\pdfoutput=1

\documentclass[11pt]{article}

\usepackage{acl}

\usepackage{times}
\usepackage{latexsym}

\usepackage[T1]{fontenc}

\usepackage[utf8]{inputenc}

\usepackage{microtype}
\usepackage{amsmath}
\usepackage{amssymb}
\usepackage{booktabs}
\usepackage{xspace}

\usepackage{inconsolata}

\usepackage{graphicx}

\newcommand{\ie}{\emph{i.e.,}\xspace}
\newcommand{\eg}{\emph{e.g.,}\xspace}

\newcommand{\SeaRefuse}{\textsc{SeaRefuse}\xspace}
\newcommand{\FreshQAParallel}{\textsc{FreshQAParallel}\xspace}
\newcommand{\TrueFalseMultiLang}{\textsc{TrueFalseMultiLang}\xspace}

\newcommand{\SeaRefuseH}{\textsc{SeaRefuse-H}\xspace}
\newcommand{\SeaRefuseG}{\textsc{SeaRefuse-G}\xspace}

\title{Analyzing LLMs' Knowledge Boundary Cognition Across Languages Through the Lens of Internal Representations}

\author{
\textbf{Chenghao Xiao}$^{1,2,}$\footnotemark[2] \ \ 
        \textbf{Hou Pong Chan}$^{1,}$\footnotemark[3]\ \ 
        \textbf{Hao Zhang}$^{1,}$\footnotemark[3] \ \
        \textbf{Mahani Aljunied}$^1$ \ \
        \\
        \textbf{Lidong Bing}$^1$ \ \
        \textbf{Noura Al Moubayed}$^2$ \ \ 
        \textbf{Yu Rong}$^{1,3}$ \\  
  $^1$ DAMO Academy, Alibaba Group
  \\
  $^2$ Department of Computer Science, Durham University
  \\
  $^3$ Hupan Lab, Hangzhou, China \\
  {\tt \{xiaochenghao.xch,houpong.chan,hz.hhea2e\}@alibaba-inc.com}
}

\begin{document}
\maketitle

\renewcommand{\thefootnote}{\fnsymbol{footnote}}
\footnotetext[2]{Work done during internship at Alibaba DAMO Academy.}
\footnotetext[3]{Corresponding authors.}
\renewcommand*{\thefootnote}{\arabic{footnote}}

\begin{abstract}
While understanding the knowledge boundaries of LLMs is crucial to prevent hallucination, research on the knowledge boundaries of LLMs has predominantly focused on English. 
In this work, we present the first study to analyze how LLMs recognize knowledge boundaries across different languages by probing their internal representations when processing known and unknown questions in multiple languages. Our empirical studies reveal three key findings: 1) LLMs' perceptions of knowledge boundaries are encoded in the middle to middle-upper layers across different languages. 2) Language differences in knowledge boundary perception follow a linear structure, which motivates our proposal of a training-free alignment method that effectively transfers knowledge boundary perception ability across languages, thereby helping reduce hallucination risk in low-resource languages; 3) Fine-tuning on bilingual question pair translation further enhances LLMs' recognition of knowledge boundaries across languages. Given the absence of standard testbeds for cross-lingual knowledge boundary analysis, we construct a multilingual evaluation suite comprising three representative types of knowledge boundary data. Our code and datasets are publicly available at \url{https://github.com/DAMO-NLP-SG/LLM-Multilingual-Knowledge-Boundaries}.
\end{abstract}

\section{Introduction}

The rapid advancement of large language models (LLMs) has revolutionized their capacity to store and utilize knowledge across languages~\cite{grattafiori2024llama3herdmodels,qwen2025qwen25technicalreport}. Understanding the knowledge boundaries of Large Language Models (LLMs) is critical, as LLMs tend to hallucinate when attempting to answer questions beyond their knowledge. Yet, research on knowledge boundaries of LLMs has predominantly focused on English~\cite{azaria-mitchell-2023-internal,marks2024geometrytruthemergentlinear,li2024knowledgeboundarylargelanguage,DBLP:conf/emnlp/CheangCW0LS0C23}. Misaligned knowledge boundaries between languages can lead to inconsistent and unsafe outputs in cross-lingual applications. Therefore, it is crucial to determine whether the knowledge boundary perceptions of LLMs observed in English can be similarly identified in other languages or transferred across them. 

\begin{figure}[t]
  \includegraphics[trim={0.7cm 0.3cm 0.6cm 0.4cm},clip,width=\columnwidth]{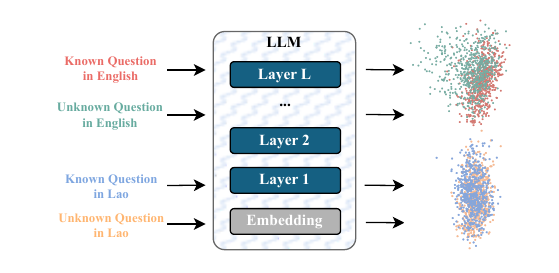}
  \caption{Our goal is to analyze LLMs' cognition of knowledge boundaries across different languages by inspecting their representations of knowledge boundaries in multiple languages. The right-hand side of the figure illustrates the representations when an LLM encodes known and unknown questions in English and Lao.}
  \label{fig:intro-figure}
\end{figure}

To fill this gap, we are pioneering the investigation into \textit{how LLMs perceive and encode knowledge boundaries across languages}, as illustrated in Figure~\ref{fig:intro-figure}. Through probing the representations of LLMs, our work reveals novel structural geometry, training-free transfer methods, and training methods to jointly enhance cross-lingual awareness. 

We begin by applying layer-wise probing on parallel multilingual questions with true/false premises to examine \textbf{how LLMs perceive knowledge boundary across languages and layers}. Initial experiments (\S\ref{sec: probing-main}) show that 1) the cognition of knowledge boundaries is encoded in the middle to mid-upper layers of LLMs. 2) The representation subspaces of different languages converge into a unified knowledge space, in which probes exhibit the best inter-language transferability. 3) Low-resource language representations provide high zero-shot transferability to high-resource language representations, but not vice versa.

Motivated by the above observations, we further explore \textbf{whether a specific structure exists in the geometry of multilingual knowledge boundary representations, allowing training-free alignment methods to transfer abilities between languages}. Experiments (\S\ref{sec: training-free}) show that language difference for knowledge boundary is encoded in a linear structure, and training-free alignment methods, such as mean-shifting and linear projection, can effectively transfer the knowledge boundary perception across languages. Notably, linear projection largely closes the performance gap between in-distribution (ID) and out-of-distribution (OOD) language representations, demonstrating improved cross-lingual transferability. We also find that projecting high-resource language representations onto low-resource subspaces removes extraneous noise encoded by the inherent high dimensionality of high-resource languages. As a result, compared to the low-resource language representation themselves, probes trained on low-resource languages achieve better performance on these projected high-resource language representations than on low-resource language representation themselves, revealing a ``weak-to-strong generalization'' pattern.

Building on the success of training-free alignment and the distinctive ``\textit{weak-to-strong generalization}'' pattern, we are intrigued by \textbf{if fine-tuning on certain languages can further refine their knowledge boundary perception ability, and generalize this improvement to other languages?} As prior work~\cite{zhang-etal-2024-getting} demonstrates that fine-tuning LLMs solely on question translation data can improve their cross-lingual performance in various downstream tasks like sentiment analysis, we investigate whether such fine-tuning can enhance LLMs' cognition of knowledge boundaries across different languages. Our experiments show that SFT using only question translation data effectively improves LLMs' perception of their knowledge boundaries across languages. We further observe a latent safeguard mechanism where the primary language representation is enhanced when the model is fine-tuned on non-dominant language pairs consisting of unanswerable questions.

To address the current lack of standard testbeds for evaluating the generalization of knowledge boundary cognition across languages, we constructed three types of multilingual knowledge boundary datasets, including: 1) questions with true/false premises; 2) entity-centric answerable/unanswerable questions; and 3) true/false statements. For true/false-premise questions, we first contribute an augmented version of \textsc{FreshQA} \cite{vu2023freshllms} by flipping the premise of each question, and creating its corresponding multilingual version. For entity-centric answerable/unanswerable questions, we create a multilingual QA dataset using an entity-swapping framework, where the content is written in authentic language for each target language, based on a diverse array of real and fictional entities. We also translate the widely used true/false statement dataset in \cite{azaria-mitchell-2023-internal} into 8 languages.

To our knowledge, we are the first to comprehensively study the generalization of LLM knowledge boundary awareness across languages. In summary, our contributions are three-fold: 1) We present the first comprehensive analysis of how LLMs' cognition of knowledge boundaries transfers across languages. 2) We develop a multilingual evaluation suite comprising three representative types of knowledge boundary data, providing valuable resources for both evaluating and enhancing LLMs' perception of knowledge boundaries across languages. 3) Our training-free transfer methods significantly reduce the performance gap in probing across languages, demonstrating effectiveness as an additional signal in LLM generation that substantially reduces the risk of hallucination.

\section{Related Work}

\begin{table*}[t]
    \small
    \centering
    \begin{tabular}{lccc}
    \toprule
         Dataset & Languages & \# train samples & \# test samples \\
    \midrule
        \FreshQAParallel & en, zh, vi, th, id, ms, km, lo & - & 9,600 (1,200 per lang.)\\
        \SeaRefuse & en, zh, id, th, vi & 64,075 ($\sim$10-15k per lang.) &  6,000 (1,200 per lang.) \\
        \TrueFalseMultiLang & en, es, de, it, pt, fr, id, th & - & 48,680 (6,085 per lang.) \\
    \bottomrule
    \end{tabular}
    \caption{Statistics of the three datasets in our constructed evaluation suite.}
    \label{tab: dataset info}
\end{table*}

\subsection{Knowledge Boundaries of LLMs}

While knowledge boundaries are crucial for mitigating hallucinations and unsafe generation, the concept remains underdefined \cite{li2024knowledgeboundarylargelanguage}. From a generative perspective, \citet{yin-etal-2024-benchmarking} classify LLM's knowledge into prompt-agnostic knowledge, prompt-sensitive knowledge, and unknown knowledge. Complementarily, representations-based studies focus on whether LLMs' knowledge boundary perception is reflected in their internal states, through probing LLM's representations with true/false statements and analyzing their geometric patterns \cite{azaria-mitchell-2023-internal,marks2024geometrytruthemergentlinear,bürger2024truthuniversalrobustdetection}. However, most work in this line is limited in true/false statements, and no work has systematically investigated multilingual knowledge boundaries, except \citet{bürger2024truthuniversalrobustdetection} briefly showed that probes trained on English show certain generalization to German statements. In this work, we show that most inter-language generalization gaps can be closed with training-free subspace alignment.

\subsection{Multilingualism in LLMs}

Multilingual large language models increasingly support diverse languages through larger-scale training \cite{qwen2025qwen25technicalreport,grattafiori2024llama3herdmodels} and targeted efforts towards multilinguality  \cite{workshop2023bloom176bparameteropenaccessmultilingual,üstün2024ayamodelinstructionfinetuned,DBLP:journals/corr/seallms3}. Studies identify language-specific neurons in multilingual large language models and steer LLMs' behaviors by perturbing such neurons \cite{tang-etal-2024-language,zhao2024largelanguagemodelshandle}, revealing sparse language-specific neurons and ``anchor language'' processing in middle layers \cite{zhao2024largelanguagemodelshandle}. \citet{mu-etal-2024-revealing} link parallel multilingual inputs during inference to inhibit neurons and precise neuron activation. Cross-lingual improvement patterns emerge during fine-tuning: \citet{zhang-etal-2024-getting} find that question translation fine-tuning boosts multilingual performance, even for languages that are not directly fine-tuned on. In this work, we provide more concrete evidence to this method from a representation perspective.

\section{Datasets}
We construct a multilingual evaluation suite to analyze how LLMs generalize their knowledge boundary cognition across languages, comprising three datasets: \FreshQAParallel, \SeaRefuse, and \TrueFalseMultiLang. These cover three types of knowledge boundary data: questions with true or false premises, entity-centric answerable/unanswerable questions, and true/false statements. Table~\ref{tab: dataset info} summarizes the dataset statistics.

\paragraph{\FreshQAParallel}
Our \FreshQAParallel dataset extends \textsc{FreshQA}~\cite{vu2023freshllms} by creating parallel true/false premise questions. Human annotators are asked to manually invert the premise type of each question, \eg converting ``When did Google release ChatGPT?'' (false-premise) to ``When did OpenAI release ChatGPT?'' (true-premise). We highlight the difficulty of converting true-premise questions to false-premise due to larger search space, and provide more details of the annotation in Appendix~\ref{sec:FreshQAParallel_details}. We further apply GPT-4o to translate the dataset into 7 languages, which are then quality-checked by linguists in the team.

\paragraph{\SeaRefuse}
We propose the \SeaRefuse Benchmark, which is composed of unanswerable questions about non-existing entities and answerable questions in English, Chinese, Indonesian, Thai, and Vietnamese. We devise an entity-swapping approach and a generation-based approach to construct questions about fake entities. The entity swapping approach constructs unanswerable questions by swapping the named entities in questions from open-source QA datasets into non-existent entities. The questions form \SeaRefuseH training/test set. Each question in the test set is verified by linguists. The generation-based approach relies on GPT-4o to synthesize unanswerable questions, resulting in the \SeaRefuseG test set. More details are in Appendix~\ref{sec:SeaRefuse_details}. 

\paragraph{\TrueFalseMultiLang} 
We apply GPT-4o to translate the \textsc{TrueFalse} dataset~\cite{azaria-mitchell-2023-internal}, a dataset of true and false statements on six different topics, into 7 languages (es, de, it, pt, fr, id, th) that form the intersection that SOTA LLMs~\cite{qwen2025qwen25technicalreport,grattafiori2024llama3herdmodels} claim to support. The quality of the translation is verified by linguists.

\section{Knowledge Boundary Probing}
\label{sec: probing-main}

We first systematically analyze our primary question: \textit{can we probe the representations of language models to detect knowledge boundaries, and what patterns emerge across languages and layers?}

\begin{figure*}
    \centering
    \includegraphics[trim={0.3cm 0.3cm 0.3cm 0.2cm},clip,width=0.95\linewidth]{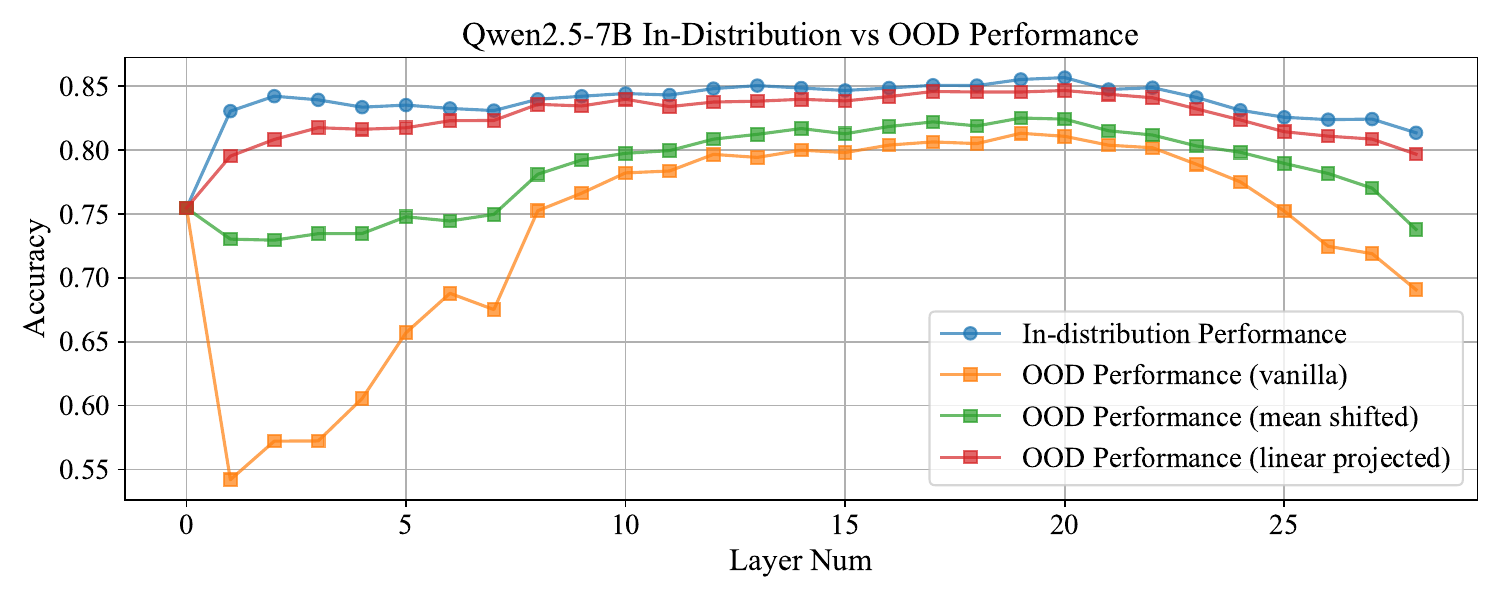}
    \vspace{-2mm}
    \caption{In-distribution and OOD performance of layer-wise probes trained on \texttt{Qwen2.5-7b} representations, across different methods. All scores are averaged across all languages.}
    \label{fig: transferability across layers (main)}
\end{figure*}

\paragraph{Experiment Settings.}
We utilize \texttt{Qwen2.5-7B} and \texttt{Llama-3.1-8B} in our main experiments and further study the scaling with \texttt{Qwen2.5-14B} and \texttt{Qwen2.5-72B}. For each model, we train $k \times m$ in-distribution linear classifiers $ f: \mathbb{R}^{d} \to \mathcal{C} $, using the last token representations of questions $\mathbf{E} \in \mathbb{R}^{n \times d} $ for each language, where $k$ is the number of layers, $m$ is the number of languages, $n$ is the number of questions and $d$ the embedding dimensionality. Each layer- and language-specific classifier (trained on in-distribution data) is evaluated zero-shot on all other languages. We use \textbf{accuracy} as the main metric for evaluating probing models. We average per-language probe performance in each layer (\eg an 8-language setup yields 64 scores per layer). 

\paragraph{RQ1.1: \textit{Do probes transfer cross-lingually, and what's the pattern across layers, languages, and models?}}

\begin{figure}[t]
    \centering
    \includegraphics[width=0.78\linewidth]{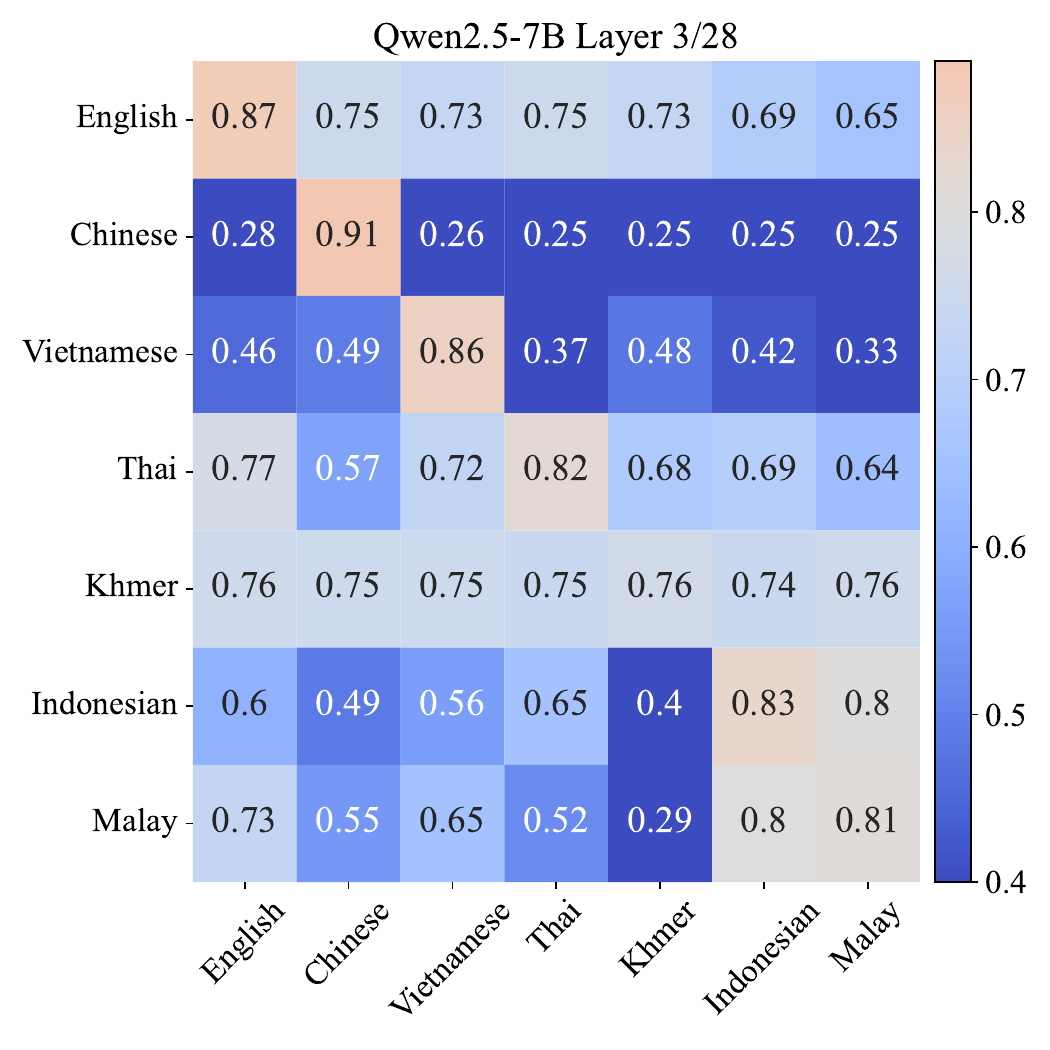}\\
    \includegraphics[width=0.78\linewidth]{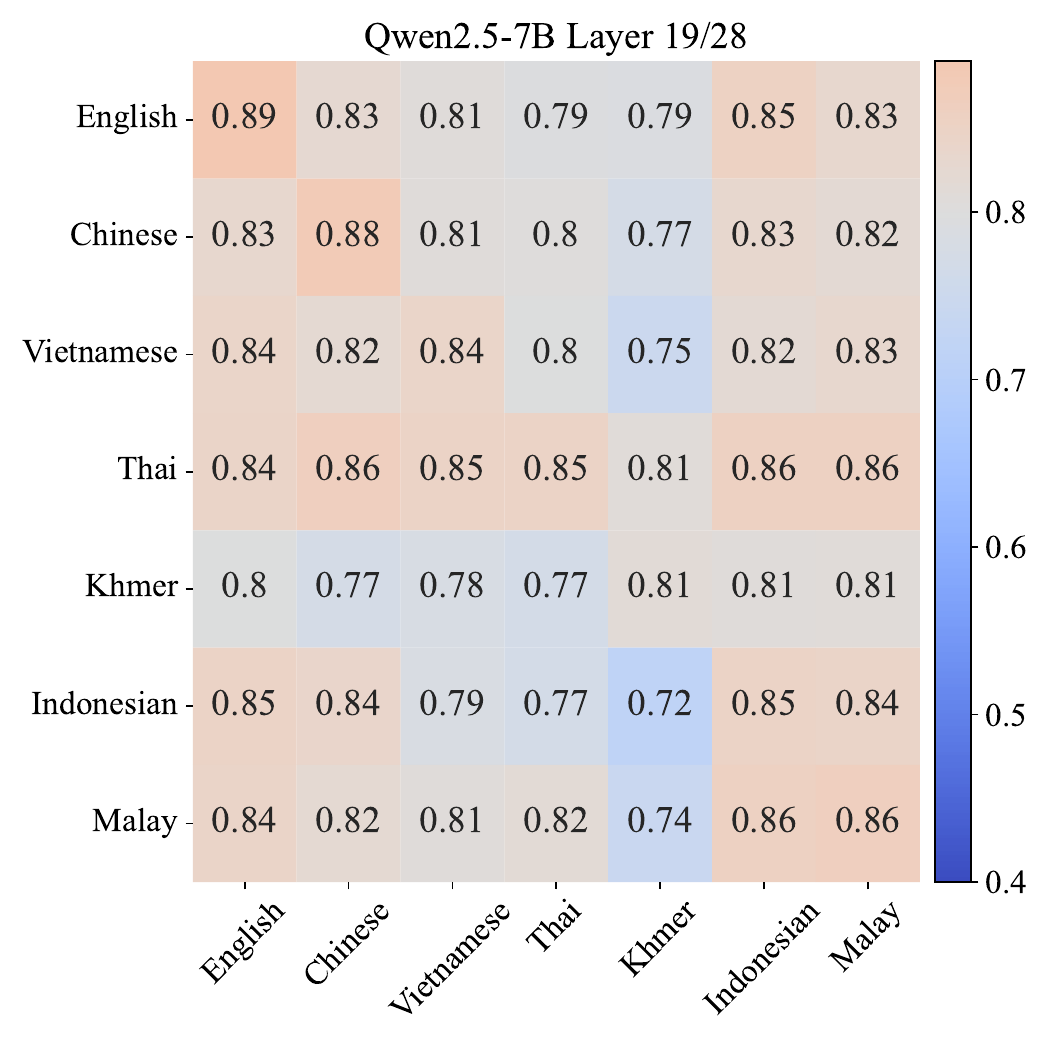}
    \vspace{-3mm}
    \caption{The performance of probing models on the 3\textsuperscript{rd} and 19\textsuperscript{th} layers of \texttt{Qwen2.5-7B} with different training and testing language combinations. The y-axis and x-axis represent the languages used for training and inference of the probing models, respectively.
    }
    \label{fig: vanilla; transferability matrix}
    \vspace{-0.1in}
\end{figure}

\paragraph{Layers.} Figure~\ref{fig: vanilla; transferability matrix} visualizes the transferability matrices of probing model on \texttt{Qwen2.5-7B}'s 3\textsuperscript{rd} and 19\textsuperscript{th} layers, which exhibit the worst and best average transferability across languages, respectively. Figure~\ref{fig: transferability across layers (main)} reveals a significant gap between in-distribution (\textcolor{blue}{blue line}) and OOD (\textcolor{orange}{orange line}) performance, which is most pronounced in the bottom layers. In these layers, language models begin incorporating language-specific static embeddings and processing within corresponding language subspaces before converging to a unified knowledge representation space shared across languages in the middle and mid-upper layers.

\paragraph{Languages.} High and low-resource languages exhibit distinct performance and transferability patterns (Figure~\ref{fig: vanilla; transferability matrix}). 
1) \textit{High-resource languages achieve superior ID accuracy but transfer poorly to low-resource languages.}
For instance, all languages provide the lowest transferability to Khmer, \eg English$\to$Khmer drops from 89\% to 79\% even in the most transferable layer.
2) \textit{Mid-resource languages offer modest transferability to both high- and similar-resource languages}, such as Thai$\to$Chinese and Thai$\to$Indonesian in the 19th layer.
3) \textit{Low-resource languages, like Khmer, show the best relative transferability}, with consistent performance across all layers and languages.
These patterns suggest that discriminative features for low-resource languages exist within high-resource representations but not vice versa.
More evidences are provided in the \textit{weak-to-strong generalization} section of \S\ref{subsec: training-free findings}.

\paragraph{Models.} 
The above findings hold across model families (\texttt{Llama}, \texttt{Qwen}), sizes (7B to 72B), and variants (base/instruct). We refer to the Appendix for full results across models. Representations show subpar cross-lingual transferability in the bottom layers. The transferability peaks in middle (\texttt{Llama}) and mid-upper layers (\texttt{Qwen}), then diverges again in final layers.

Our results offer more concrete and direct evidence for the 3-phase hypothesis in \citet{zhao2024largelanguagemodelshandle}, where language models first process input in the source language, then ``think'' in English or other anchor languages (such as English and Chinese for \texttt{Qwen}), and finally switch back to the source language for generation. 

\paragraph{RQ1.2: \textit{Do language models fail to express their knowledge boundary multilingually, even when the boundary is stored in their representations?}}

\noindent The promising probing performance demonstrates that LLMs internally encode knowledge boundary awareness, such as identifying false-premise questions. But have they fully utilized the awareness in generation? We compare models' performance on FreshQA's false-premise subset with and without hints suggesting the question may be false-premise. As shown in Table~\ref{tab:performance with hints}, providing premise hints significantly improves the performance of probes across languages, with the largest gain of 
$24.49\%$ for \texttt{Qwen2.5-7B-Instruct} on Vietnamese and $17.69\%$ for \texttt{Qwen2.5-72B-Instruct} on Malay.
These results indicate that while models possess discriminative representations of knowledge boundaries, they may underutilize the information during generation. This highlights the value of training multilingual knowledge boundary probes in practical applications, like detecting false-premise on the fly and prompting re-generation with corrected hints.

\begin{table}[t]
    \centering
    \resizebox{.47\textwidth}{!}{\begin{tabular}{lcccccccc}
         \toprule
         & en & zh & vi & th & km & id & ms & lo \\ 
         \midrule
         \multicolumn{8}{l}{\textit{Qwen2.5-7B-Instruct}}\\
         Baseline & 30.61 & 36.05 & 19.73 & 19.73 & 8.16 & 22.45 & 19.05 & 0.68 \\ 
         FP-Hinted & 41.50 & 45.58 & 44.22 & 32.65 & 11.56 & 38.10 & 37.41 & 2.04 \\ 
         \midrule
         \multicolumn{8}{l}{\textit{Qwen2.5-72B-Instruct}}\\
         Baseline & 58.50 & 60.54 & 61.90 & 55.10 & 33.33 & 59.18 & 55.78 & 31.29 \\ 
         FP-Hinted & 72.11 & 70.75 & 68.03 & 67.35 & 44.90 & 72.79 & 73.47 & 38.10 \\ 
         \bottomrule
    \end{tabular}}
    \caption{Performance comparison between models without and with hints suggesting the question may be false-premise (Baseline vs. FP-Hinted).}
    \label{tab:performance with hints}
    \vspace{-0.1in}
\end{table}

\section{Training-free Transfer}
\label{sec: training-free}

Given the distinct patterns across languages, a natural question is \textit{whether knowledge boundary perception capabilities can be transferred between languages?} In this section, we investigate the geometric structure in the knowledge boundary representations and explore training-free alignment methods to exploit such structures.

\subsection{Subspace Geometry}

\noindent\textbf{RQ2.1: \textit{Does a linear structure exist in the geometry of knowledge boundary representations across languages?}}

We investigate whether an LLM's perception towards knowledge boundaries is encoded in language-neutral ways. Using embeddings from layers that yield the best cross-language transferability (\eg the 19\textsuperscript{th} layer of \texttt{Qwen2.5-7B}), three LDA classifiers are trained on encodings from the concatenated \TrueFalseMultiLang dataset across all languages. The classifiers are trained using three different label sets: 1) language labels, 2) the Cartesian product of the domain set and truth/falseness, and 3) binary truth/falseness labels.

\begin{figure*}
    \centering
    \includegraphics[width=1\linewidth]{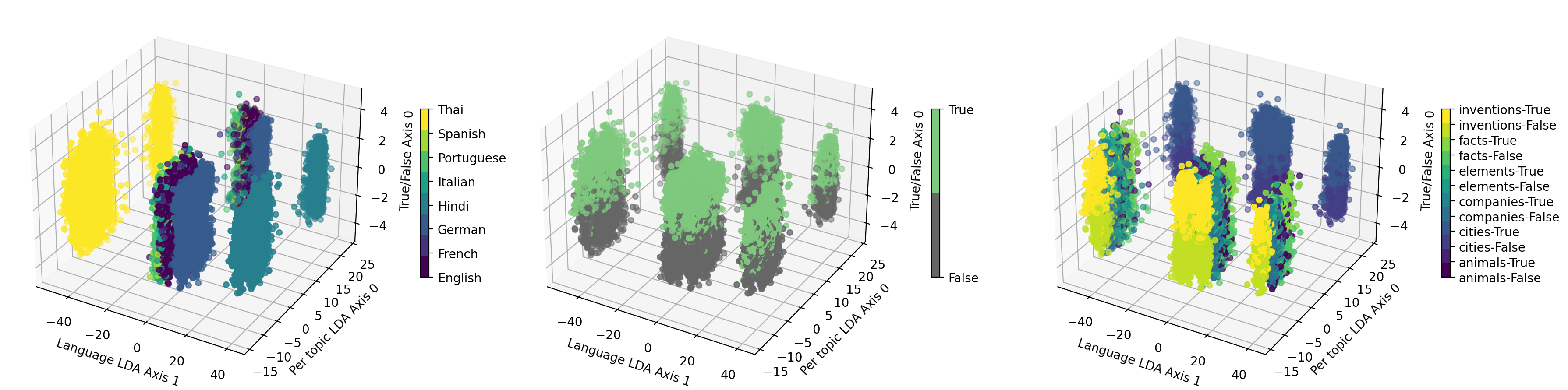}
    \caption{We project the 19\textsuperscript{th} layer of \texttt{Qwen2.5-7B} onto a linear subspace where the x-axis encodes languages, the y-axis encodes topic-aware true/falsity, and the z-axis encodes binary truth/falsity. We visualize the same scatter plot with three different ground-truth label sets annotated respectively.}
    \label{fig:geometry}
\end{figure*}

Figure~\ref{fig:geometry} visualizes embeddings projected onto axes taken from the three LDA subspaces. Annotated with the three label sets, the results reveal: 1) the model represents questions in a language-neutral way, presenting a parallel structure in the projected axes (Figure~\ref{fig:geometry}, left); 2) truth/falsity is encoded within the model (Figure~\ref{fig:geometry}, middle); and 3) true/false statements can be separated in a topic-agnostic way by a near-horizontal hyperplane (Figure~\ref{fig:geometry}, right). Similar patterns emerge for entity-aware binary answerability in the \SeaRefuse dataset (Appendix~\ref{appendix: lda sea refuse}, Figure~\ref{fig: lda sea refuse}).

\subsection{Subspace Projections}

\noindent\textbf{RQ2.2: \textit{Can we transfer the abilities to perceive knowledge boundaries between languages by linearly aligning their representation space without training?}}

\noindent Motivated by the language neutrality in representing true/falsity, we further explore the effectiveness of training-free embedding subspace alignment methods, \ie \textbf{Mean Shifting} and \textbf{Linear Projection}, to exploit linear structures and align language subspaces.

\textbf{Mean Shifting.} 
We denote the in-distribution (source language) and out-of-distribution (target language) training sets as $\mathbf{X}_{\text{in}}^{\text{train}} \in \mathbb{R}^{n \times d} $ and $ \mathbf{X}{_\text{ood}}^{\text{train}} \in \mathbb{R}^{n \times d} $, respectively, where $ n $ is the number of parallel samples and $ d $ is the feature space dimensionality. Prior work~\cite{chang-etal-2022-geometry,xu-etal-2023-language-representation} reveals that language differences are mainly encoded by language subspace representation means in multilingual encoder models. We assess whether mean-shifting can serve as an effective baseline for LLMs. This method aims to adjust target language embeddings so their mean aligns with the source language embeddings. The mean features for both distributions can be computed as:
$\boldsymbol{\mu}_{\text{in}} = \frac{1}{n} \sum_{i=1}^{n} \mathbf{x}_{\text{in}, i},  \boldsymbol{\mu}_{\text{ood}} = \frac{1}{n} \sum_{i=1}^{n} \mathbf{x}_{\text{ood}, i}$,
where $ \mathbf{x}_{\text{in}, i} $ and $ \mathbf{x}_{\text{ood}, i} $ are the embeddings of $ i $-th sample in $ \mathbf{X}_{\text{in}}^{\text{train}} $ and $ \mathbf{X}_{\text{ood}}^{\text{train}} $, respectively, from which the difference between the two mean vectors can be attained:
$\Delta \boldsymbol{\mu} = \boldsymbol{\mu}_{\text{in}} - \boldsymbol{\mu}_{\text{ood}}.$

This difference vector $ \Delta \boldsymbol{\mu} $ represents the translation needed to align the mean of the OOD subspace with that of the in-distribution subspace, which is then applied to the OOD test set $ \mathbf{X}_{\text{ood}}^{\text{test}} \in \mathbb{R}^{p \times d} $ to obtain the shifted test set $ \mathbf{X}_{\text{shifted}}^{\text{test}} $:
$\mathbf{X}_{\text{shifted}}^{\text{test}} = \mathbf{X}_{\text{ood}}^{\text{test}} + \Delta \boldsymbol{\mu}.$ Appendix~\ref{appendix: non-parallel mean-shifting} shows that language means can also be approximated with non-parallel corpus.

\textbf{Linear Projection}
Using the same notation of $ \mathbf{X}_{\text{in}}^{\text{train}} \in \mathbb{R}^{n \times d} $ and $ \mathbf{X}_{\text{ood}}^{\text{train}} \in \mathbb{R}^{n \times d} $, the linear transformation $ \mathbf{W} \in \mathbb{R}^{d \times d} $ that projects vectors from the target language subspace to the source language subspace is estimated by solving the least squares problem:
$\mathbf{W} = \arg \min_{\mathbf{W}} \| \mathbf{X}_{\text{in}} - \mathbf{X}_{\text{ood}}^{\text{train}} \mathbf{W} \|_F^2,$
where $ \| \cdot \|_F $ denotes the Frobenius norm. The solution for $ \mathbf{W} $ is computed using the Moore-Penrose pseudo-inverse:
$    \mathbf{W} = \mathbf{X}_{\text{ood}}^{\text{train}+} \mathbf{X}_{\text{in}}$, where \( \mathbf{X}_{\text{ood}}^{\text{train}+} = \mathbf{V} \mathbf{\Sigma}^+ \mathbf{U}^\top \) is derived from the Singular Value Decomposition (SVD) \( \mathbf{X}_{\text{ood}}^{\text{train}} = \mathbf{U} \mathbf{\Sigma} \mathbf{V}^\top \). Here, \( \mathbf{\Sigma}^+ \) replaces non-zero singular values \( \sigma_i \) with \( 1/\sigma_i \). For full-rank \( \mathbf{X}_{\text{ood}}^{\text{train}} \), it reduces to \( \mathbf{W} = (\mathbf{X}_{\text{ood}}^{\text{train}\top} \mathbf{X}_{\text{ood}}^{\text{train}})^{-1} \mathbf{X}_{\text{ood}}^{\text{train}\top} \mathbf{X}_{\text{in}} \). Given that our representation matrices mostly don't have full column rank (\eg the number of questions in FreshQA is far smaller than embedding dimensionality), SVD is used to preserve numerical stability.
 
Once $ \mathbf{W} $ is estimated, it can be used to project the out-of-distribution test set $ \mathbf{X}_{\text{ood}}^{\text{test}} \in \mathbb{R}^{p \times d} $ into the in-distribution subspace, resulting in the shifted test set $ \mathbf{X}_{\text{shifted}}^{\text{test}} $:
$\mathbf{X}_{\text{shifted}}^{\text{test}} = \mathbf{X}_{\text{ood}}^{\text{test}} \mathbf{W}$. 
For scalability, inter-language projection matrices can be precomputed by sampling $\mathbf{X}{\text{\small ood}}$ and $\mathbf{X}{\text{\small in}}$ from larger parallel corpora $\mathbb{C}_{\text{ood}}$ and $\mathbb{C}_{\text{in}}$.

\begin{figure*}
    \centering
    \includegraphics[width=1.0\linewidth]{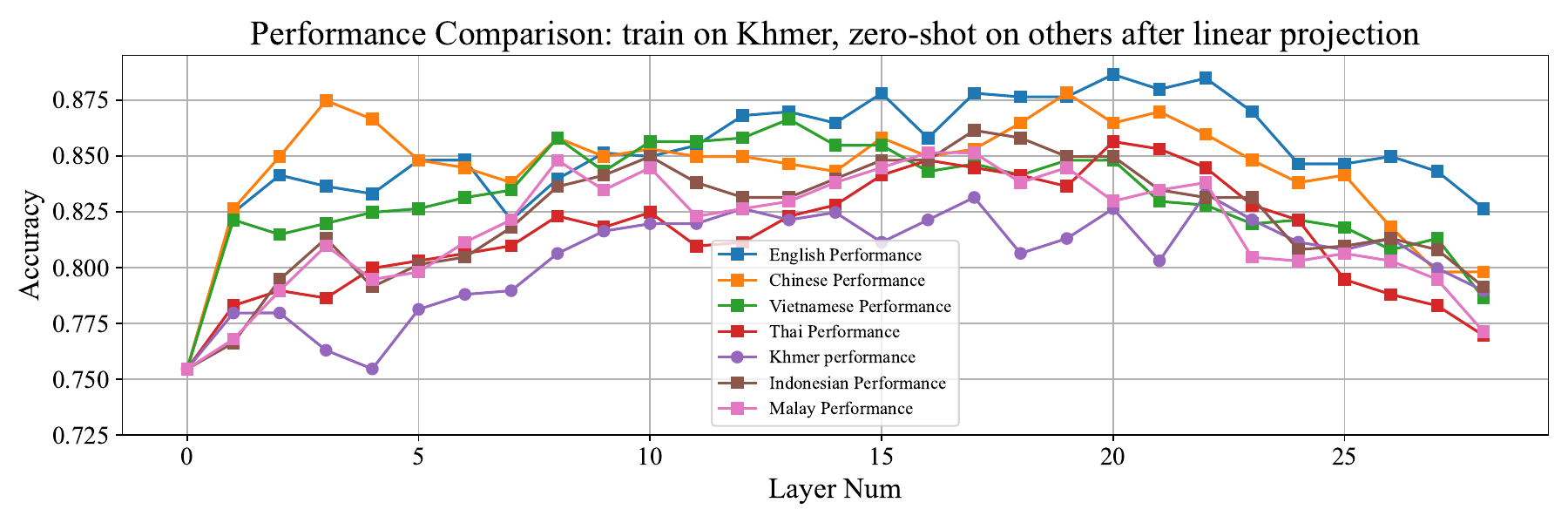}
    \caption{Performance of layer-wise probes trained on Khmer for different languages post-projection. We use linear projection to project each language's representations onto the subspace of Khmer. 
    It is observed that Khmer probes perform better on OOD languages than on the Khmer language itself. }
    \label{fig:khmer transfer qwen base}
    \vspace{-0.1in}
\end{figure*}

\subsection{Findings}
\label{subsec: training-free findings}

\paragraph{Mean Shifting largely improves transferability.} As shown in Figure~\ref{fig: transferability across layers (main)}, mean shifting (\textcolor{green}{green line}) substantially enhances transferability across layers compared to vanilla representations, indicating that language difference is largely encoded in subspace mean. However, the remaining gap compared to in-distribution performance suggests that additional factors, such as orientation and dimensional magnitudes (\ie anisotropic stretching in each dimension), also affect the subspace differences.

\paragraph{Linear Projection nearly matches in-distribution performance.} In Figure~\ref{fig: transferability across layers (main)}, linear projection (\textcolor{red}{red line}) nearly closes the gap with in-distribution accuracy. The linear transformation, learned from language-pair training sets, robustly captures geometric relationships rather than overfitting to example-specific features.

The gap between mean-shifted and linearly-projected representations highlights the additional complex mapping needed to align language subspaces. Meanwhile, the promising cross-lingual transferability of probes on projected representations highlights their applicability for generation. In practice, one can pretrain probes on all languages and precompute linear projection for every language pair. In inference time, a lightweight language detector can be applied to identify the inferred language, and use the linear projection $\mathbf{W}_{\text{j→k}}$ to transform language $j$'s representations into language $k$'s subspace (the language that transfers best to $j$) to use the optimal probes.

Using \texttt{Qwen2.5-14B} as an example, the best performance for Thai ($88.31\%$) and Malay ($88.64\%$) is achieved by applying the 30\textsuperscript{th}-layer Chinese probe after linearly projecting onto the Chinese subspace, rather than using their probes.
Similarly, Khmer ($88.15\%$) and Lao ($86.64\%$) perform best with the 29\textsuperscript{th}-layer Malay and 32\textsuperscript{nd}-layer Thai probes, respectively. This high$\to$mid, mid$\to$low transferability trend showcases the relative dominance of language representations due to levels of language resources in the pretraining of LLMs.

\paragraph{A weak-to-strong generalization pattern.} We observe an intriguing pattern: probes trained on low-resource languages (\eg Khmer, Lao) perform better on other languages after post-projection than on their own (\eg the Khmer example in Figure~\ref{fig:khmer transfer qwen base}).

\begin{figure*}
    \centering
    \includegraphics[width=0.95\linewidth]{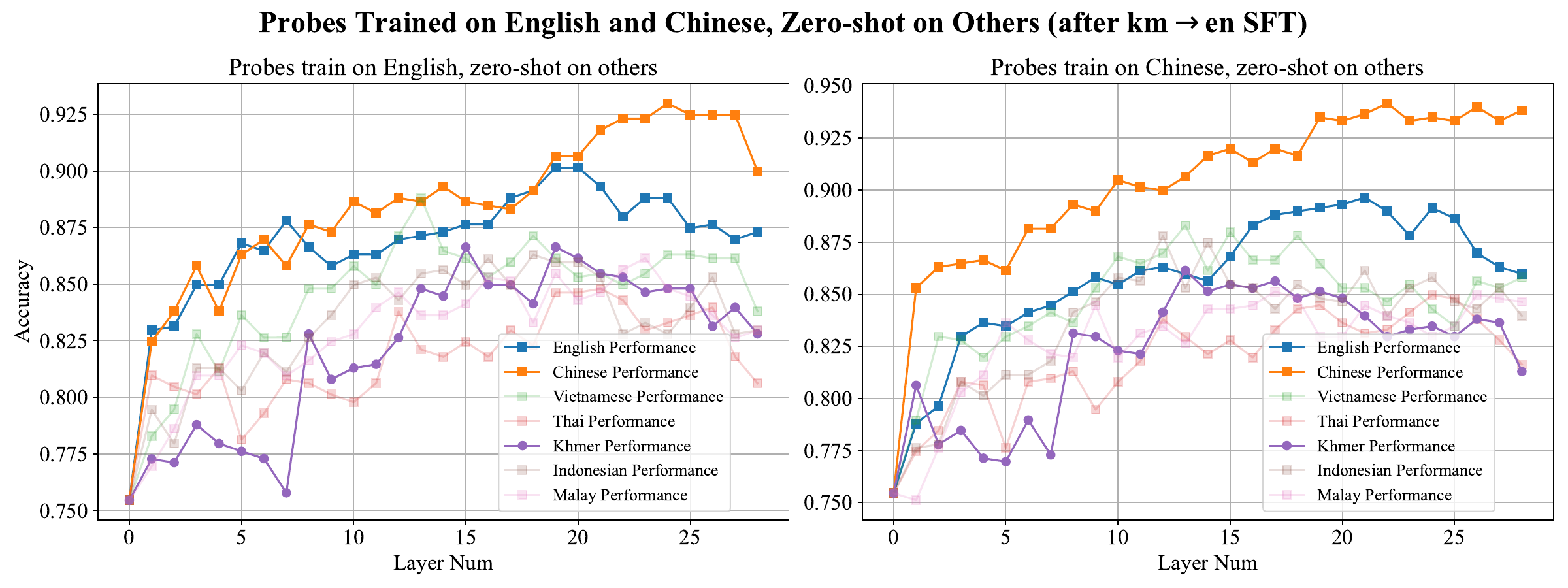}
    \vspace{-3mm}
    \caption{After SFT'ed on Khmer-English translation pairs, \texttt{Qwen2.5-7B} exhibits unexpected enhancement in Chinese representations, which otherwise would achieve $\sim0.90$ or lower accuracy across probes.}
    \label{fig:pattern-2}
\end{figure*}

We hypothesize that projecting high-resource language representations onto low-resource subspaces acts as a denoising/regularization step, removing language-specific nuances such as syntactic specifics. Table~\ref{tab: effective dimensionality} provides empirical evidence using \FreshQAParallel. By dividing the examples into $80\%$-$20\%$ train-test splits and ensuring each original question with its flipped-premise counterpart in the same split, we compute the SVD of the centered test set representations $\mathbf{X} \in \mathbb{R}^{n \times d}$, yielding singular values $\{\sigma_i\}_{i=1}^{n}$, and approximate effective dimensionality by $
\min \left\{ k \in \mathbb{N} \mid \sum_{i=1}^k \sigma_i^2 \geq 0.95 \cdot \sum_{i=1}^n \sigma_i^2 \right\}
$, reflecting the compactness of task-relevant information in the representation space.\footnote{The effective dimensionality is approximated by the minimum number of principal components required to explain 95\% of the total variance} We also compute the participation ratio (PR) as $
\frac{\left( \sum_{i=1}^n \sigma_i^2 \right)^2}{\sum_{i=1}^n \sigma_i^4}
$.\footnote{Participation ratio (PR) denotes the effective number of dimensions contributing significantly to the total variance.} Next, we project languages onto other subspaces via $\mathbf{W}$, computed using $\mathbf{X}_{\text{lang1}}^{\text{train}} \in \mathbb{R}^{4n \times d}$ and $\mathbf{X}_{\text{lang2}}^{\text{train}} \in \mathbb{R}^{4n \times d}$, and recalculate these metrics.

\begin{table}[]
    \centering
    \resizebox{.45\textwidth}{!}{\begin{tabular}{ccccc}
         \toprule
         & \multicolumn{2}{c}{Effective Dimensionality} &
         \multicolumn{2}{c}{Participation Ratio}\\ 
         \cmidrule{2-5}
          & Original & Projected & Original & Projected \\ 
         \midrule
         Khmer & 103 & 97  & 15.93 & 18.07\\
         English & 116 & 87  & 26.26 & 19.29 \\ 
         \bottomrule
    \end{tabular}}
    \caption{Effective Dimensionality and Participation Ratios: original subspace and after English and Khmer projected onto each other's subspace.}
    \label{tab: effective dimensionality}
\end{table}

Table~\ref{tab: effective dimensionality} compares the results in English (high-resource) and Khmer (low-resource). English's higher intrinsic effective dimensionality suggests it encodes more features, potentially language-specific and task-irrelevant details. In contrast, Khmer's subspace is more compact and task-focused, likely due to limited training data that forces LLMs to prioritize generalizable features during pretraining. However, projecting English onto Khmer's subspace reduces its effective dimensionality ($116\to87$), stripping away noise or redundant features. English's higher PR indicates diffuse variance across many directions, whereas Khmer's lower PR reflects tighter feature alignment with the tasks. 
English $\to$ Khmer subspace reduces PR ($26.26\to19.29$), aligning closer to Khmer’s compactness. Khmer → English subspace increases PR ($15.93\to18.07$), indicating a loss of focus. Thus, low-resource subspaces act as inductive biases for cross-lingual tasks by filtering noise and preserving core semantics, while high-resource subspaces risk carrying superfluous details that degrade transfer.

\section{Training-based Enhancement}
\label{sec: training SFT}
Having shown the promising performance of aligning representations across language subspaces and the weak-to-strong generalization in language pairs, we further investigate \textit{whether training on specific languages can enhance performance that generalizes to others.}
Prior research~\cite{zhang-etal-2024-getting} shows that fine-tuning question translation data, even without ground-truth answers, enhances LLM performance across languages in various downstream tasks. Inspired by it, we further ask: 

\paragraph{RQ3: Can finetuning on question translation data enhance the knowledge boundary cognition ability of LLMs across languages?}

\paragraph{Experiment Settings.}
We fine-tune \texttt{Llama-3.1} and \texttt{Qwen2.5} (from 7B to 72B) via SFT on question translation pairs (\eg English-Lao), which are constructed using the English subset of \SeaRefuse, translated by GPT-4o and verified by linguists.
We measure the performance post-SFT using the same protocols in \S\ref{sec: probing-main} and \S\ref{sec: training-free}. 

We conduct experiments on all model sizes with both Full-parameter fine-tuning and LoRA. For LoRA, we use a rank 32 and alpha 64. We use a learning rate of $1e-5$ for full fine-tuning and $1e-4$ for LoRA. A global batch size of $512$ is maintained for all settings through different gradient accumulation setups. Qwen2.5-72B is trained on 4 nodes of 32 A800 80G GPUs in total with DeepSpeed ZeRO-3 for full-parameter training.

\paragraph{SFT on the bilingual translation pairs consistently enhances cross-lingual knowledge boundary cognition.} 
Probes trained on \texttt{Qwen2.5-7B} and \texttt{Qwen2.5-72B} show consistent gains across languages after Khmer$\to$English translation fine-tuning (see Figures~\ref{fig: 7b before after}, \ref{fig: 72b before after} in Appendix~\ref{appendix: full translation results}), with \texttt{Qwen2.5-7B} achieving a $+2.3\%$ average gain in the best layer, yielding a $88\%$ average accuracy in classifying true/false premises. This improvement holds across language families and bilingual pair selections: fine-tuning \texttt{Llama-3.1-8B} on Khmer$\to$Thai or Lao$\to$English pairs enhances upper-layer performance by $>10\%$ (Figures~\ref{fig: layers-llama-3.1-8B-km-th}, \ref{fig: layers-llama-3.1-8B-lo-en}), indicating the upper-layer ``expression spaces'' across languages has been better aligned.

\paragraph{A \textit{self-defense} mechanism.}
Fine-tuning on low-resource$\to$English translation pairs unexpectedly enhances the model's strongest inherent language. For instance, \texttt{Qwen2.5} fine-tuned on Khmer-English pairs exhibits emergent alignment favoring Chinese representations (Figure~\ref{fig:pattern-2}), significantly improving performance with probes trained in other languages (full results in Figure~\ref{fig:pattern-2-full_7B}). We hypothesize that processing unanswerable questions in low-resource languages (\eg Khmer) activates latent safety mechanisms linked to the model's primary languages, akin to a ``self-defense'' reflex where boundary awareness in weaker languages reinforces safeguards in stronger ones.

\section{Conclusion}
We present the first comprehensive study on how LLMs perceive knowledge boundaries across languages through a systematic analysis of their internal representations. Our findings reveal that knowledge boundary cognition is encoded in middle to middle-upper layers across languages, following a linear structural pattern that enables effective cross-lingual transfer. Motivated by this, we propose a training-free alignment method and targeted fine-tuning with question translation data to transfer knowledge boundary perception ability across languages. Our multilingual evaluation suite provides a valuable resource for future research. These insights not only advance our understanding of LLMs' cross-lingual knowledge boundary perception but also offer practical strategies for improving transfer to low-resource languages.

\section*{Limitations}
While we extensively study probing representations of LLMs' perception of questions, it is of interest to investigate how these representations evolve in the generation process. For instance, given the paradigm shifts to explicit long chain-of-thoughts in the generation process, it is possible that LLMs reach the ``aha moment'' in the reasoning process when they become aware that the question reaches its knowledge boundaries, where this representation becomes more linearly separable to the probes that classify this perception. However, we note that it is very difficult to locate the information when the transformation takes place and extract the corresponding representations. We leave this interesting line of exploration to future work. 

\bibliography{acl}

\appendix

\section{Language-neutrality of Entity-aware binary Answerability}
\label{appendix: lda sea refuse}

Having shown the linear structure of true/false statement representations across languages, we visualize the \SeaRefuse benchmark dataset we construct in this paper here. We concatenate the training set of \SeaRefuse of all 5 languages, yielding a representation set of over 120k for training LDA classifiers.

\begin{figure*}
    \centering
    \includegraphics[width=\linewidth]{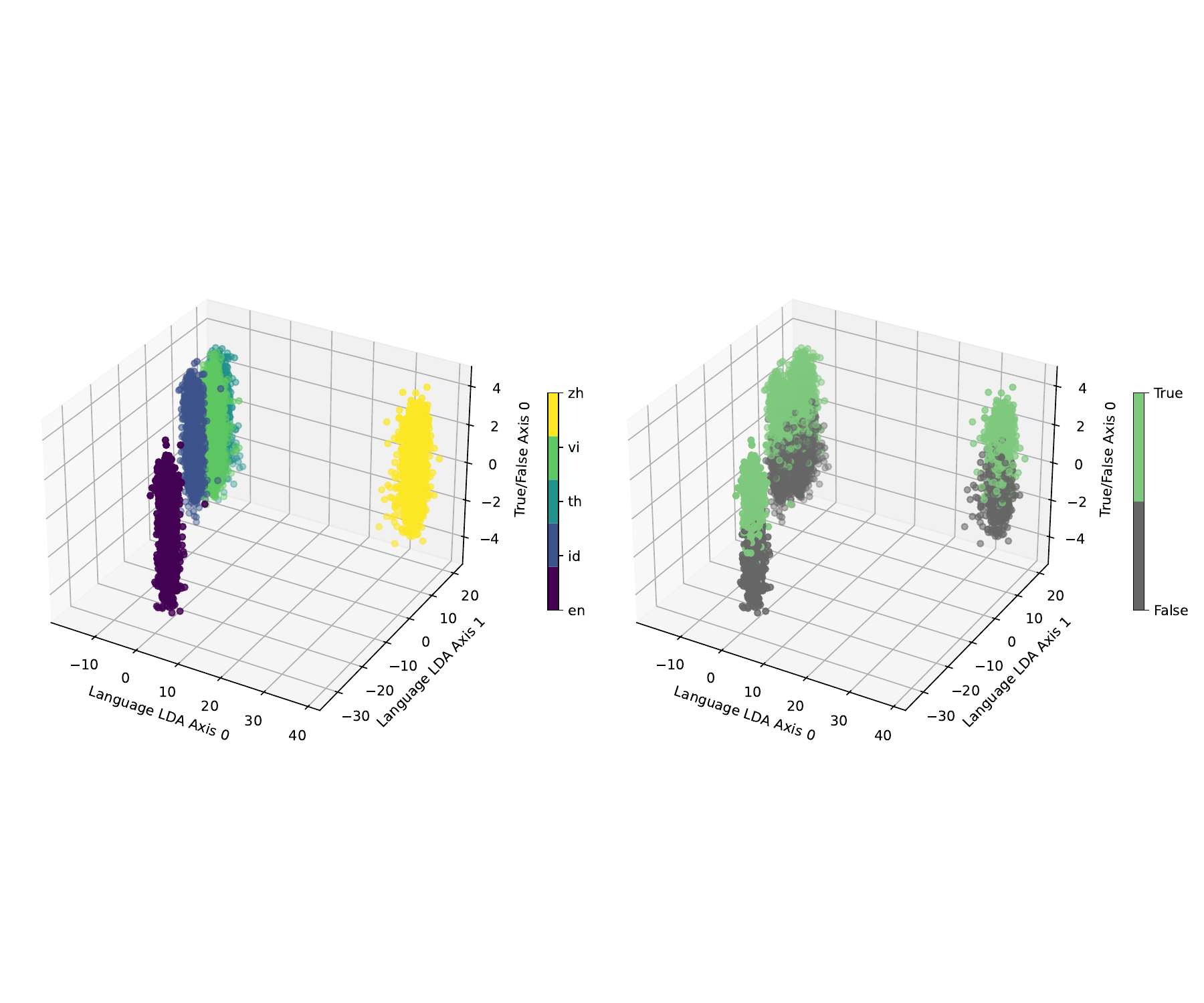}
    \caption{Projecting Qwen-2.5-7B's 19\textsuperscript{th} layer representations onto two languages axes (x,y), and one answerability axis (z) - determined by whether the entities in the question exist in real world.}
    \label{fig: lda sea refuse}
\end{figure*}

Due to the lack of topic and domain information as in the true/false statement datasets, we train 2 LDA classifiers with the concatenated \SeaRefuse representations of all languages, with 1) language labels. 2) binary existence labels of entities (exist/non-exist). In Figure~\ref{fig: lda sea refuse}, we project the representations onto a subspace where the horizontal axes (x,y) are from the language LDA subspace, and the vertical axis (z) is from the exist/non-existing (thus answerable/unanswerable) subspace.

\section{Transferability across Layers Full Results}

We include full results of transferability across layers brought by our training-free methods, across model families and sizes. \eg Figure~\ref{fig: layers-llama-3.1-8B} presents results of Llama-3.1-8B.

\section{Transferability Pattern Change after SFT'ed on Question Pairs}
\label{appendix: full translation results}

\begin{figure*}
    \centering
    \includegraphics[width=1.0\linewidth]{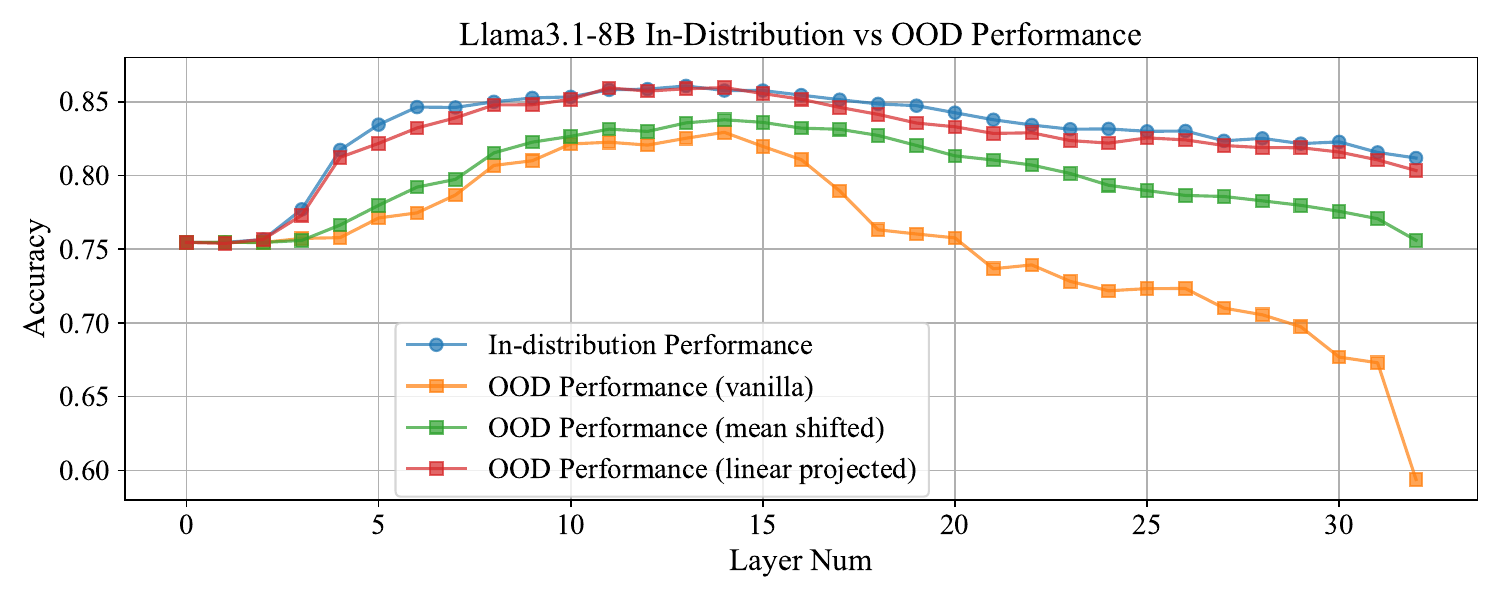}
    \caption{Llama-3.1-8B training-free alignment methods performance.}
    \label{fig: layers-llama-3.1-8B}
\end{figure*}

\begin{figure*}
    \centering
    \includegraphics[width=1.0\linewidth]{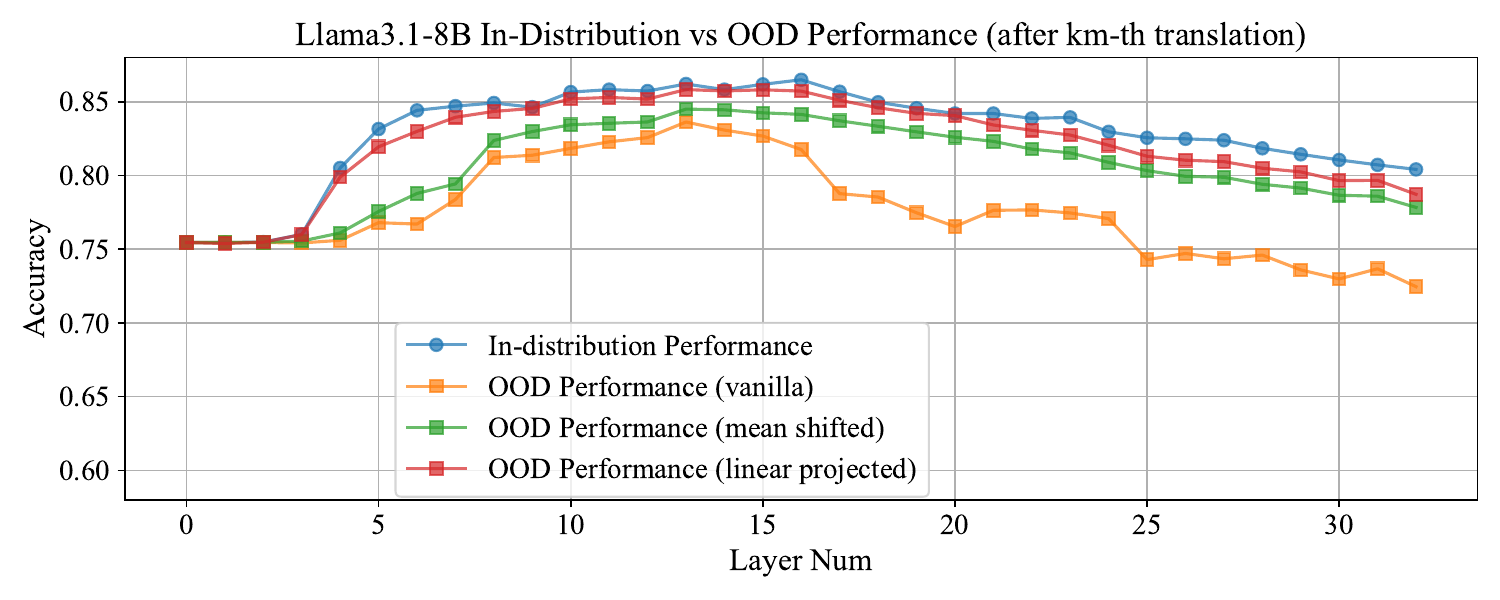}
    \caption{Llama-3.1-8B training-free alignment methods performance, after the model has been SFT'ed on Khmer-Thai translation.}
    \label{fig: layers-llama-3.1-8B-km-th}
\end{figure*}

\begin{figure*}
    \centering
    \includegraphics[width=1.0\linewidth]{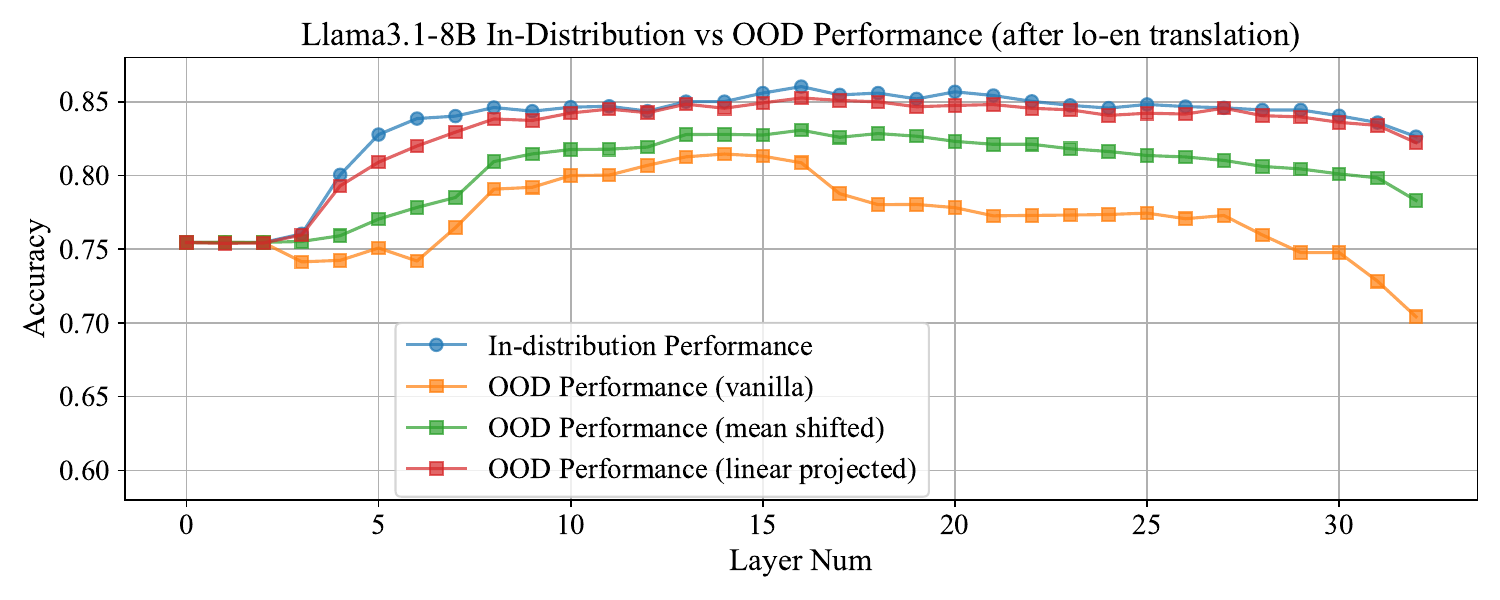}
    \caption{Llama-3.1-8B training-free alignment methods performance, after the model has been SFT'ed on Lao-English translation.}
    \label{fig: layers-llama-3.1-8B-lo-en}
\end{figure*}

Figure~\ref{fig: layers-llama-3.1-8B-km-th}, and ~\ref{fig: layers-llama-3.1-8B-lo-en} present Llama-3.1-8B performance on FreshQA, after the model has been fine-tuned on Lao-English, Khmer-Thai SeaRefuse question translation pairs without answerability indications, respectively. As seen, the last layers show a drastic improvement over Figure~\ref{fig: layers-llama-3.1-8B} (before training), showing an internal alignment brought to across languages, beyond the bilingual pairs involved in training.

\section{Pattern 2 Full Visualization}
\begin{figure*}
    \centering
    \includegraphics[width=0.9\linewidth]{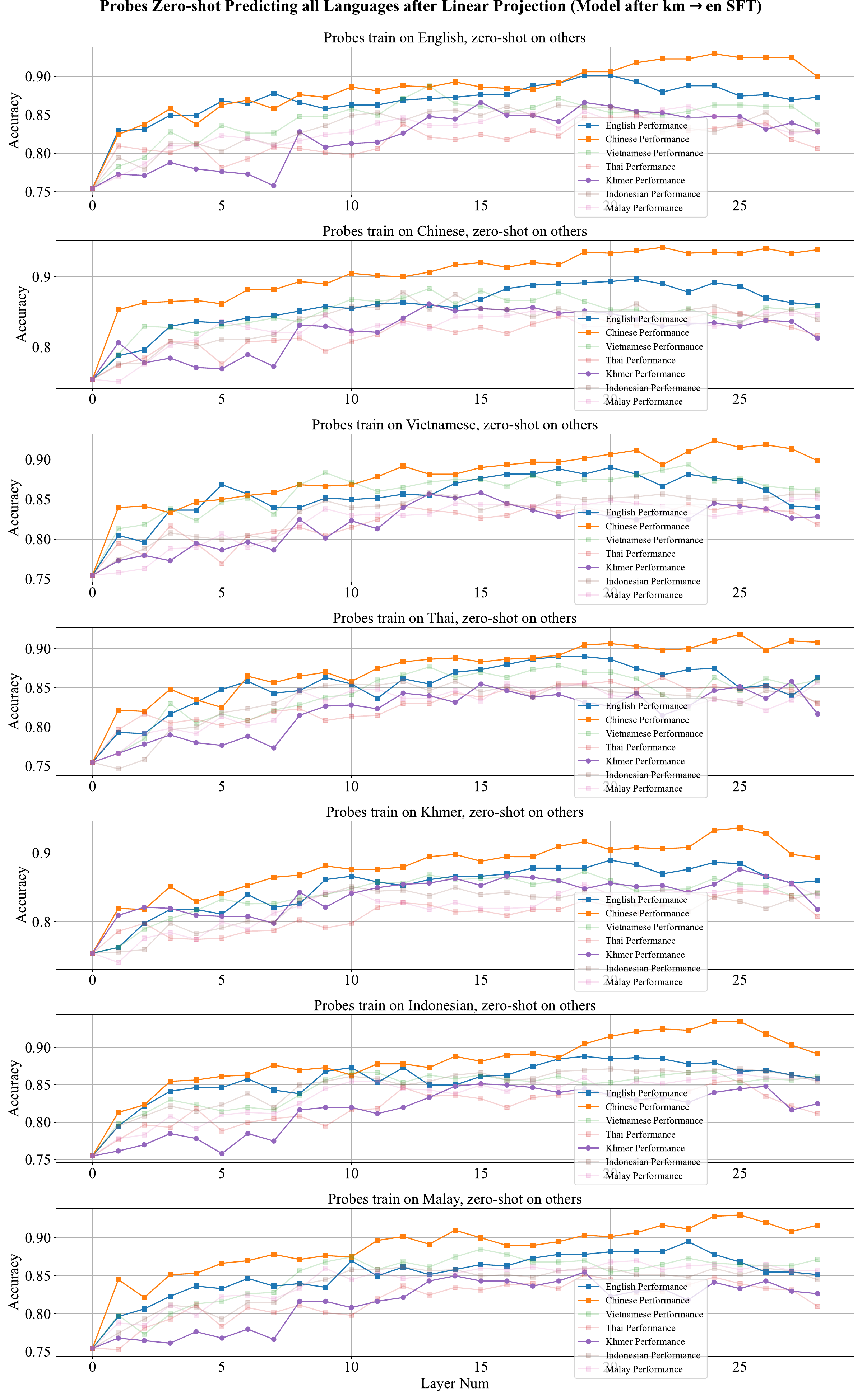}
    \caption{Full transferability results for probes trained on all languages, after \texttt{Qwen2.5-7B} has been fine-tuned on Khmer-English translation.}
    \label{fig:pattern-2-full_7B}
\end{figure*}

Figure~\ref{fig:pattern-2-full_7B} visualizes the self-defense pattern of Qwen2.5 across all language probes.


\section{Details of Data Construction of \FreshQAParallel}\label{sec:FreshQAParallel_details}

Although flipping a false-premise question to true-premise could be simpler by referring to search engines or reasoning based on the groundtruth answers about the original questions in FreshQA, flipping a true-premise question to false-premise is more difficult, given the larger search space (many ways to make it false-premise). We note that, the flipped question must be talking about the same topic with the original question; without getting too creative (e.g., GPT-4o would frequently fail to do this task as it generates too fictional entities such as transforming QS World Ranking to Hogwarts Ranking); and not grounded in the futuristic property relative to model's training cutoff time (e.g., Who is the US president in 2024 is not deemed a false-premise question to a model trained in 2023, although it's unanswerable without retrieval-augmented generation; However ``Who was the US president in 2035'' is false-premise, which assumes a future date is in the past). 

Below are some more difficult examples we communicate to the human annotators and our linguist team:

Example 1.a. “How many children does Leonardo DiCaprio have?”
Questions like these are not false-premise questions. Even though Leonardo DiCaprio doesn’t have children, this question doesn’t assume that he has, as the answer to this question can be 0, and thus answerable. 

Example 1.b. “How old is Leonardo DiCaprio’s second kid?”
This is a valid false-premise question because it has assumed a false fact (that the entity has 2 or more kids).

Similarly, 
Example 2.a. “How many championships has Chris Paul won?” (not a valid false-premise question because it doesn’t assume anything; the answer can be 0).
Example 2.b. “When did Chris Paul win the second championship?” (a valid false-premise question because it has assumed a false fact.)

For question (id 124) How many humans have landed on Mars? This is not a false-premise question because it’s answerable. Questions like “How many humans have landed on Venus?” are not valid false-premise either. In this case, we can write something like “How many humans have landed on Mars as part of the first manned mission in 2020?” that meets the requirements (making the question not answerable because non-factual). 

\section{Details of Data Construction of \SeaRefuse}\label{sec:SeaRefuse_details}
We first prompt GPT-4o to generate fake entity candidates of the types of person, location, organization, event, and work of art, respectively.
The entity swapping approach first applies the spacy\footnote{https://spacy.io} NER tool to detect named entities from human-written answerable factoid questions. Then, we randomly select a named entity from each source question and then swap the selected entity with a random entity from the non-existing entity candidates of the same type. As the detected named entities may not be accurate, we further ask linguists to verify the quality of the constructed unanswerable questions and rectify the disfluent questions, resulting in the \textsc{SeaRefuse-H} test set. This set contains 100 answerable and 100 unanswerable questions for each language.
The generation-based approach first prompts GPT-4o to write a 120-word story for a randomly selected fake entity candidate. Then we prompt GPT-4o to write four factoid questions about the selected fake entity based on the generated story, resulting in the SeaRefuse-G test set. This set includes 500 answerable and 500 unanswerable questions for each language, except for Vietnamese, which has 483 of each.

The answerable questions are collected from open source factoid QA training sets, including ParaRel~\cite{r-tuning,DBLP:journals/tacl/ElazarKRRHSG21}, SQUAD~\cite{DBLP:conf/acl/squad2}, Web Questions QA\footnote{https://github.com/brmson/dataset-factoid-webquestions}, TempQ~\cite{DBLP:conf/www/tempQ}, NLPCC-KBQA~\cite{NLPCC-KBQ16,NLPCC-KBQ17}, iapp-wiki-qa\footnote{https://github.com/iapp-technology/iapp-wiki-qa-dataset}, vi-quad~\cite{DBLP:conf/coling/vi-quad}, facqa-qa-factoid-itb~\cite{wilie2020indonlu}, and TyDi QA~\cite{tydiqa}.

\begin{table}[t]
    \centering
    \resizebox{0.47\textwidth}{!}{\begin{tabular}{lccccc}
    \toprule
         & en & id & th & vi & zh\\
    \midrule
         before & 97.25 & 95.83 & 93.25 & 95.45 & 93.58\\
         after & 97.67 & 95.92 & 93.33 & 95.20 & 95.92 \\
         gain & 0.42 & 0.09 & 0.08 & -0.25 & 2.34 \\
    \bottomrule
    \end{tabular}}
    \caption{\SeaRefuse test set performance after training on km-en translation on the \SeaRefuse English training subset.}
    \label{tab: sea-refuse test}
\end{table}

\section{Additional Results on \SeaRefuse}

Table~\ref{tab: sea-refuse test} presents the per-language probing results of on the \SeaRefuseH and \SeaRefuseG concatenated test set, before and after \texttt{Qwen2.5-7B} going through Khmer$\to$English training on the \SeaRefuse English training subset. Note that translating from Khmer to English only covers entities that would appear in English contexts. The model's generalization to the \SeaRefuse test set, which comprises entities grounded in each language, shows the effectiveness of the method. As seen in Table~\ref{tab: sea-refuse test}, the Chinese representations also achieve a large gain even without being directly trained on, in line with the safeguard mechanism of primary language we find in the main paper.

\begin{table}[t]
\centering
\resizebox{0.47\textwidth}{!}{\begin{tabular}{lcccc}
\toprule
 & layer 4 & layer 5 & layer 17 & layer 18 \\ \midrule
vanilla & 53.10 & 66.25 & 85.25 & 86.03 \\
parallel & \textbf{75.57} & \textbf{78.90} & \textbf{86.53} & \underline{86.48} \\
non-parallel & \underline{71.98} & \underline{77.85} & \underline{86.00} & \textbf{86.56} \\ \bottomrule
\end{tabular}}
\caption{Conducting mean-shifting with parallel examples and non-parallel examples.}\label{tab: non-parallel mean-shifting}
\end{table}

\begin{table}[t]
\centering
\begin{tabular}{lcc}
\toprule
 & ID avg. & OOD avg. \\
 \midrule
mean pooling & 86.12 & 79.52 \\ 
last-token pooling & \textbf{86.77} & \textbf{83.95} \\ 
\bottomrule
\end{tabular}
\caption{In-distribution and OOD performance across different pooling methods.}\label{tab: pooling ablation}
\end{table}

\section{Additional Results on \FreshQAParallel}

\begin{figure*}
    \centering
    \includegraphics[width=1.0\linewidth]{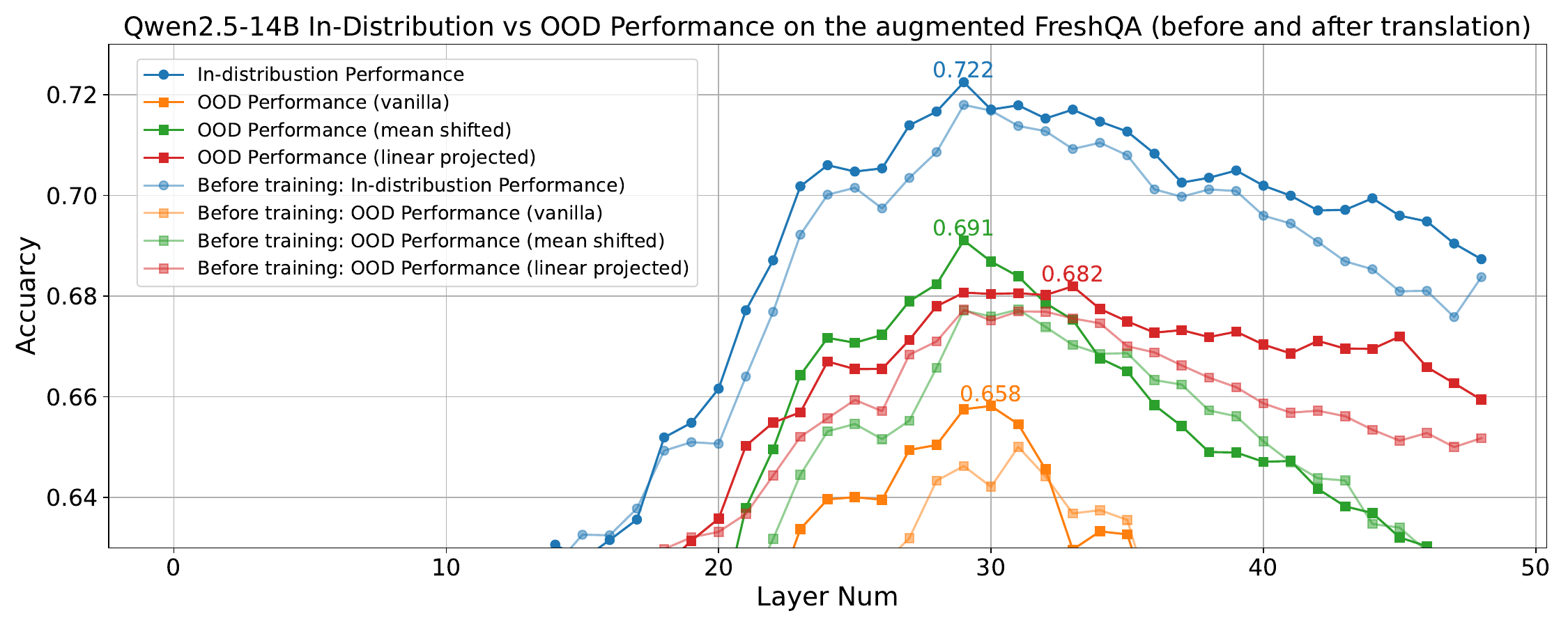}
    \caption{14B model performance on our augmented \FreshQAParallel before and after bilingual translation training on km-en question pairs.}
    \label{fig: freshqa-augment-14B}
\end{figure*}

Here, we provide additional \texttt{Qwen2.5-14B} results on our augmented \FreshQAParallel in Figure~\ref{fig: freshqa-augment-14B}. While our augmented \FreshQAParallel makes FreshQA much more difficult compared to the original version, bilingual question translation training still brings consistent gain to our mean-shifting and linear projection method, consistently outperforming vanilla zero-shot transfer.

\section{Non-parallel Approximation for Mean-shifting}
\label{appendix: non-parallel mean-shifting}
Here, we discuss whether our methods can be applied on non-parallel examples of large scales. 

For mean-shifting, larger non-parallel corpus can work as well. We conduct the following experiments with English, Thai and Chinese:
To construct non-parallel samples to compute language means, we sample $\sim$7k examples per language from our SeaRefuse training set, which are \textbf{non-parallel} as all questions are grounded in different language's contexts collected from different seed datasets. We run the same FreshQA experiment in the main paper with the three languages, where the parallel setting uses means computed on the fly with FreshQA training examples in each run ($\sim$480 pairs). As shown in Table~\ref{tab: non-parallel mean-shifting}, means computed by non-parallel samples also effectively approximate mean difference across languages, performing on-par with the parallel setting or even outperforming it in some cases (e.g., in layer 18).

In contrast, linear projection naturally requires aligned data. In this work, we show that a few hundreds of parallel data samples suffice to learn a robust projection matrix. 

\section{Pooling Methods}

In Table~\ref{tab: pooling ablation}, we ablate the effect of pooling methods, validating the choice of last-token pooling for knowledge boundary probing.

\begin{figure*}
    \centering
    \includegraphics[width=0.9\linewidth]{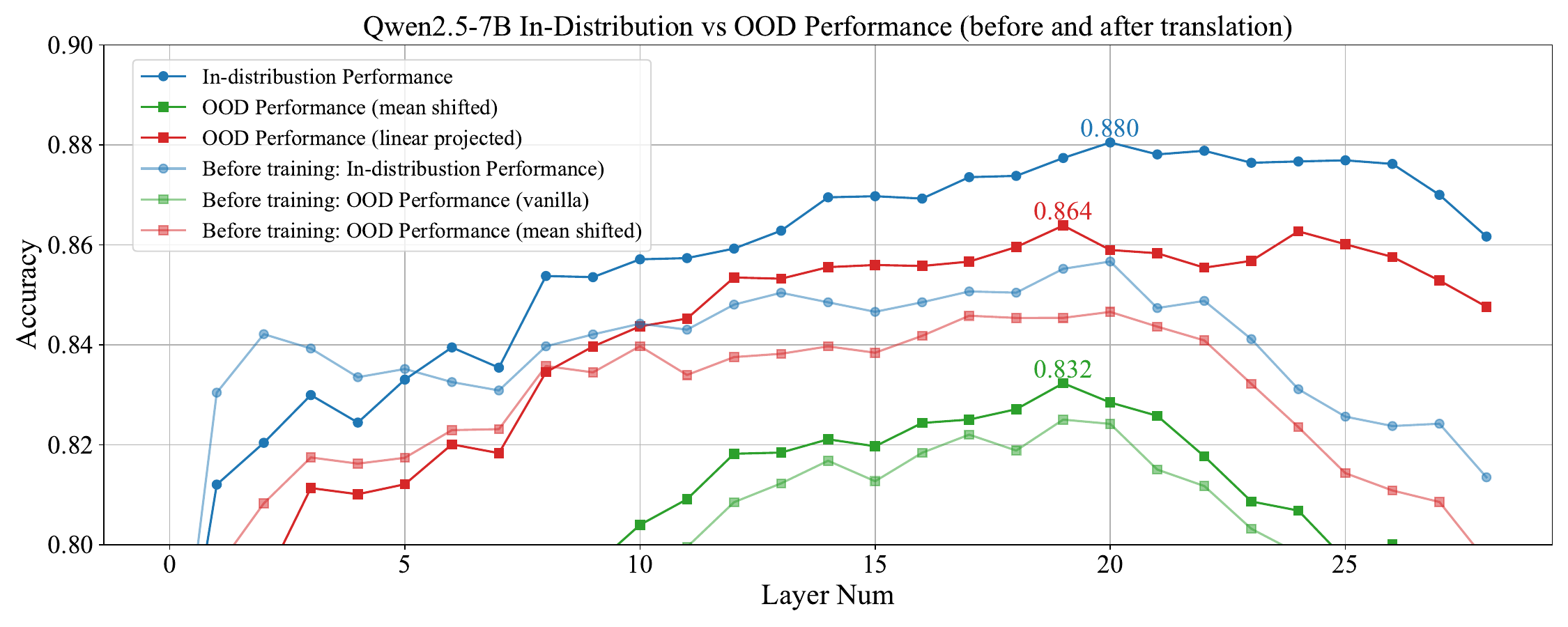}
    \caption{7B model performance on FreshQA before and after bilingual translation training on km-en question pairs. The results are averaged across all languages, aligning with the setting throughout the paper.}
    \label{fig: 7b before after}
\end{figure*}

\begin{figure*}
    \centering
    \includegraphics[width=0.9\linewidth]{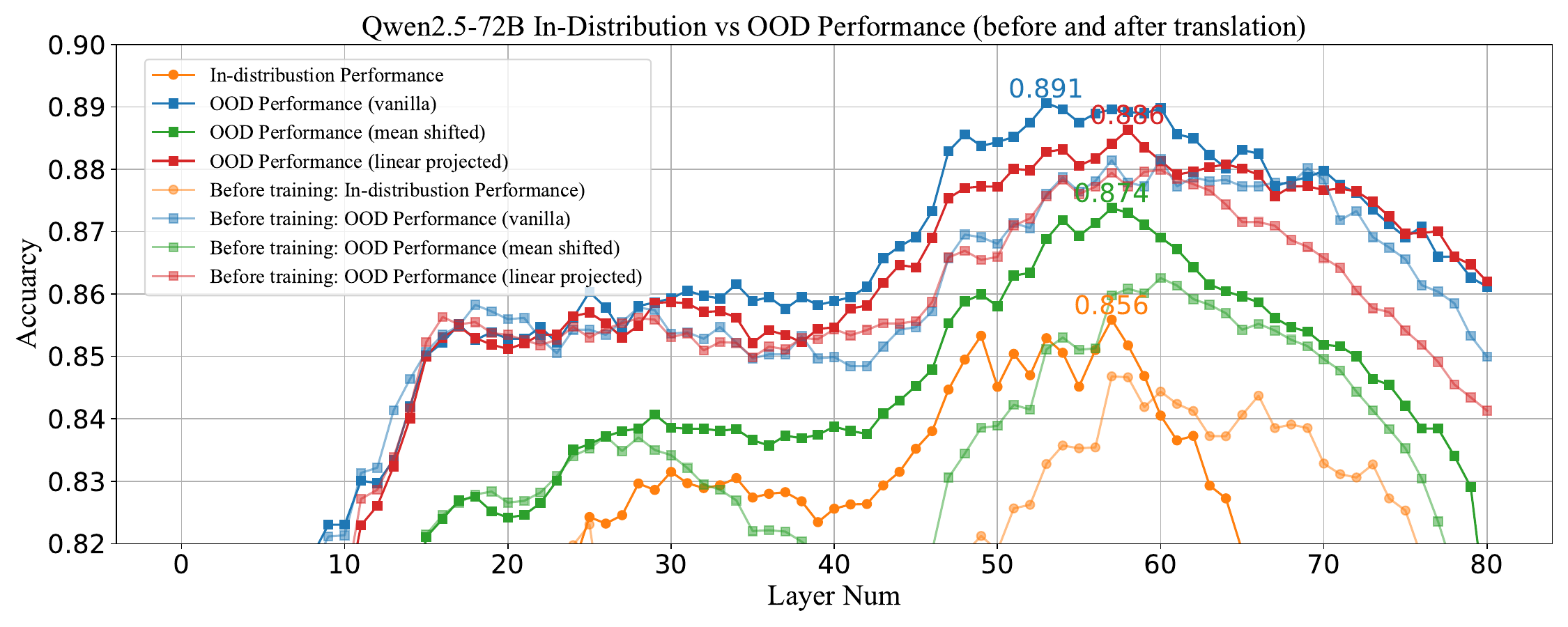}
    \caption{72B model performance on FreshQA before and after bilingual translation training on km-en question pairs. The results are averaged across all languages, aligning with the setting throughout the paper.}
    \label{fig: 72b before after}
\end{figure*}

\newpage 

\end{document}